\documentclass[lettersize,journal]{IEEEtran}
\usepackage{amsmath,amsfonts}
\usepackage{algorithmic}
\usepackage{algorithm}
\usepackage{array}
\usepackage[caption=false,font=normalsize,labelfont=sf,textfont=sf]{subfig}
\usepackage{textcomp}
\usepackage{stfloats}
\usepackage{url}
\usepackage{verbatim}
\usepackage{graphicx}
\usepackage{cite}
\usepackage{amssymb}
\usepackage{diagbox}
\usepackage{makecell}
\usepackage{color}
\usepackage{booktabs}
\usepackage{multirow}
\usepackage{rotating}
\usepackage{fancyhdr}
\fancypagestyle{firstpage}
{
    \pagestyle{fancy}
    \fancyhead[LO]{\textbf{This work has been submitted to the IEEE for possible publication. Copyright may be transferred without notice, after which this version may no longer be accessible.}}
}

\begin{document}

\title{A Sea-Land Clutter Classification Framework for Over-the-Horizon-Radar Based on Weighted Loss Semi-supervised GAN}

\author{Xiaoxuan Zhang, Zengfu Wang, Kun Lu, Quan Pan, and Yang Li
\thanks{This work was in part supported by the National Natural Science Foundation of China~(grant no. 61790552, U21B2008) and Natural Science Basic Research Plan in Shaanxi Province of China (2021JM-06). }
\thanks{Xiaoxuan Zhang, Zengfu Wang, Quan Pan, Yang Li are with the School of Automation, Northwestern Polytechnical University, and the Key Laboratory of Information Fusion Technology, Ministry of Education, Xi'an, Shaanxi, 710072, China.
Kun Lu is with Nanjing Research Institute of Electronics Technology and the Sky-Rainbow United Laboratory, Nanjing, Jiangsu,
210039, China.
E-mail: (xiaoxuanzhang@mail.nwpu.edu.cn; wangzengfu@nwpu.edu.cn;
mimimomoba@gmail.com;
quanpan@nwpu.edu.cn; liyangnpu@mail.nwpu.edu.cn). 
(Corresponding author: Zengfu Wang.)
}
}


\maketitle
\thispagestyle{firstpage}

\begin{abstract}
 Deep convolutional neural network has made great achievements in sea-land clutter classification for over-the-horizon-radar~(OTHR). 
 The premise is that a large number of labeled training samples must be provided for a sea-land clutter classifier.
 In practical engineering applications, it is relatively easy to obtain label-free sea-land clutter samples.
 However, the labeling process is extremely cumbersome and requires expertise in the field of OTHR.
To solve this problem, we propose an improved generative adversarial network, namely weighted loss semi-supervised generative adversarial network~(WL-SSGAN).
Specifically, we propose a joint feature matching loss by weighting the middle layer features of the discriminator of semi-supervised generative adversarial network.
Furthermore, we propose the weighted loss of WL-SSGAN by linearly weighting standard adversarial loss and joint feature matching loss.
The semi-supervised classification performance of WL-SSGAN is evaluated on a sea-land clutter dataset.
The experimental results show that WL-SSGAN can improve the performance of the fully supervised classifier with only a small number of labeled samples by utilizing a large number of unlabeled sea-land clutter samples.
Further, the proposed weighted loss is superior to both the adversarial loss and the feature matching loss.
Additionally, we compare WL-SSGAN with conventional semi-supervised classification methods and demonstrate that WL-SSGAN achieves the highest classification accuracy.

\end{abstract}


\section{Introduction}
\IEEEPARstart{A}{s} a crucial system for remote sensing, sky-wave over-the-horizon radar (OTHR) is widely used in military and civilian fields~\cite{geng2018target,lan2020measurement,thayaparan2019high,hu2018knowledge}.
The sea-land clutter classification of OTHR is the process of identifying whether the background clutter of each range-azimuth cell is originated from land or sea. 
Matching the derived classification results with the a prior geographic information cost-effectively provides coordinate registration parameters for target localization, 
which are important to the performance improvement of OTHR~\cite{guo2021improved, yun2003research}.

In recent years, deep learning has been successfully applied to sea-land clutter classification.
Compared with conventional classification methods~\cite{cuccoli2010sea,cacciamano2012coordinate,jin2012svm,turley2013high,holdsworth2017skywave} in which the design of feature extractor depends on human engineering, the deep learning-based sea-land clutter classification method has powerful feature extraction ability, which can automatically extract hierarchical representations from sea-land clutter data.
The extracted features are beneficial for subsequent classification tasks.
Li \emph{et al.}~\cite{li2019sea} proposed a sea-land clutter classification method with multiple hidden layers, which automatically extracts features at different levels from a large number of sea-land clutter data.
Considering the multi-scale and multi-resolution characteristics of sea-land clutter, Li \emph{et al.}~\cite{li2022cross} proposed a cross-scale sea-land clutter classification method based on algebraic multigrid.
However, as a fully supervised classification framework, the sea-land clutter methods in~\cite{li2019sea,li2022cross} rely on a large number of labeled training samples.
In general, the labeling process is extremely cumbersome and requires expertise in the field of OTHR.
Different from~\cite{li2019sea,li2022cross}, 
Zhang \emph{et al.}~\cite{zhang2022data} proposed a semi-supervised sea-land clutter data augmentation and classification method based on auxiliary classifier variational autoencoder generative adversarial network, which extracts the potential features of data from a small number of labeled sea-land clutter samples to achieve high-quality generation of samples and high-precision classification of clutter.

Inspired by game theory, Goodfellow \emph{et al.}~\cite{goodfellow2014generative} proposed generative adversarial network~(GAN), 
which is an implicit probability density generative model to learn the distribution of real data.
GAN was originally proposed for image synthesis~\cite{denton2015deep,radford2015unsupervised}. Subsequently, it has been applied to various fields, such as machine translation, image super-resolution, natural language processing, data augmentation, and domain adaptation, etc~\cite{wu2018adversarial,wang2018esrgan,yu2017seqgan,fahimi2020generative,kang2020effective}.
Since GAN is able to learn the intrinsic distribution of real data from unlabeled samples, it can also be used for semi-supervised learning.
The basic idea of GAN for semi-supervised learning is to combine the conventional fully supervised loss with the unsupervised loss of GAN, and then use it as the loss function of semi-supervised classifier to guide the update of network parameters. Up to now, various semi-supervised learning methods based on GAN have been developed. Categorical generative adversarial network~(CatGAN)~\cite{springenberg2015unsupervised} aims to learn a discriminator which distinguishes the samples into $K$ categories, instead of learning a binary discriminator. The supervised loss of the discriminator of CatGAN is the cross-entropy between the predicted conditional distribution
and the true label distribution of samples.
Semi-supervised generative adversarial network~(SSGAN)~\cite{odena2016semi} extends the output of the discriminator into $K+1$ classes, where the first $K$ classes correspond to 
real samples and the class $K+1$ corresponds to
generated samples.
On the basis of SSGAN, Salimans \emph{et al.}~\cite{salimans2016improved} pointed out that replacing standard adversarial loss with feature matching loss is potentially beneficial to obtain better semi-supervised classification performance.
Dai \emph{et al.}~\cite{dai2017good} proved that the generator and the discriminator are hard to achieve optimality simultaneously.
Triple GAN~\cite{li2017triple} improves GAN by using a ``three players'' framework with a generator, a discriminator and a classifier, which addresses
the issue that the generator and the discriminator of SSGAN can not be optimal at the same time.
Augmented bidirectional generative adversarial network~(Augmented BiGAN)~\cite{kumar2017semi} extends BiGAN to a semi-supervised framework, obtaining the tangent spaces of the image manifold by the generator.
The estimated tangents infer the desirable invariances  which can be injected into the discriminator to  improve semi-supervised classification performance.

GAN has the potential to extract deep level features from a large number of unlabeled sea-land clutter samples and then assist a small number of labeled samples for semi-supervised classification.
In~\cite{zhang2022data}, we used a small number of labeled samples for learning. 
Different from~\cite{zhang2022data}, in this work we use a large number of unlabeled samples for semi-supervised learning. 
Since feature matching loss is affected by the signal features extracted from the middle layer of the discriminator, and the randomness of signal may lead to the randomness of the features extracted from a single layer, we propose a joint feature matching loss by weighting multi-layer feature matching loss.
Moreover, standard adversarial loss is disturbed by the randomness of signal, which makes the training of SSGAN very difficult. 
To this end, 
we propose a weighted loss by linearly weighting standard adversarial loss and joint feature matching loss.
On this basis, we propose an improved SSGAN framework for sea-land clutter semi-supervised classification, 
namely weighted loss semi-supervised generative adversarial network~(WL-SSGAN).
The experimental results show that
the proposed weighted loss is superior to both the adversarial loss and the feature matching loss, 
and WL-SSGAN outperforms conventional methods on sea-land clutter semi-supervised classification. 
In summary, our main contributions are three-fold:
\begin{itemize}
\item Taking into account of the randomness of sea-land clutter, we design a weighted loss function, which first weights multi-layer feature matching loss and then weights joint feature matching loss and adversarial loss to improve the semi-supervised classification performance of sea-land clutter.
\item Based on the weighted loss, we propose a novel framework for sea-land clutter semi-supervised classification, namely WL-SSGAN, which can use a small number of labeled samples and a large number of unlabeled samples for end-to-end semi-supervised classification.
\item  The semi-supervised classification performance of WL-SSGAN is verified on a sea-land clutter dataset. The experimental results show that WL-SSGAN can improve the performance of fully supervised classifiers trained with only a small number of labeled samples.
\end{itemize}

The remainder of this paper is organized as follows.
In Section \ref{sec:Theoretical Background}, 
GAN is briefly introduced.
In Section \ref{sec:Sea-land Clutter Semi-supervised Classification Based on WL-SSGAN}, the network structure, loss function and training procedure of the proposed WL-SSGAN are described.
In Section \ref{sec:Experiment and Evaluation}, 
the sea-land clutter dataset used is introduced, and then the corresponding network design and network training are given, followed by the performance evaluation of WL-SSGAN.
Section \ref{sec:Conclusions and Future Work} draws the conclusions and highlights some future work.

\section{Background}\label{sec:Theoretical Background}
\subsection{Generative Adversarial Network}
\begin{figure}[htp]
\centering
\includegraphics[width=3.5in]{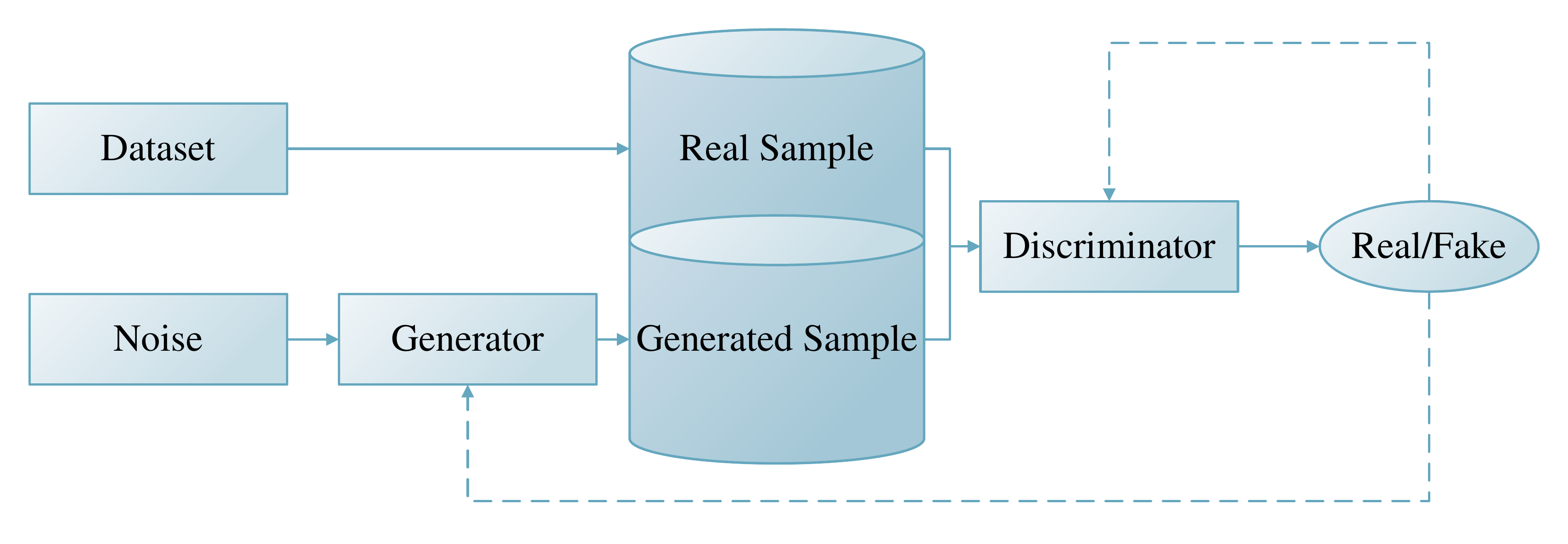}
\caption{The network structure of GAN.}
\label{fig:GAN}
\end{figure}

The network structure of the standard GAN is shown in Fig. \ref{fig:GAN}.
Specifically, GAN consists of two neural networks, a generator and a discriminator.
The role of the generator is to fool the discriminator by generating samples consistent with real-data distribution.
The role of the discriminator is to distinguish real samples in database from those generated by the generator.
The back propagation algorithm is adapted to optimize the network parameters of the generator and the discriminator alternately.
When the iteration reaches the Nash equilibrium, the adversarial network is optimal.
The generator has the ability to generate optimal samples and the discriminator cannot distinguish the source of samples.
The loss function of GAN is as follows:
\begin{equation}
\begin{aligned}
\label{eq:L_GAN}
L_{\text{GAN}}(D, G) = \min_{G} \max_{D} \, & \mathbb{E}_{x\sim p_{\text{data}}(x)}\log D(x)\\
& + \mathbb{E}_{z\sim p_{z}(z)}[1 - \log D(G(z))],
\end{aligned}
\end{equation}
where $x$ is real sample and follows the real-data distribution $p_{\text{data}}(x)$, $z$ is noise and follows the a prior distribution $p_{z}(z)$, $D(x)$ represents the probability that $x$ is a real sample, $D(G(z))$ represents the probability that $D(z)$ is a generated sample, and $\mathbb{E}$ is expectation operator. The loss function of the discriminator is $\max\limits_{D}\mathbb{E}_{x\sim p_{\text{data}}(x)}\log D(x)+\mathbb{E}_{z\sim p_{z}(z)}[1 - \log D(G(z))]$.
The loss function of the generator is $\min\limits_{G}\mathbb{E}_{z\sim p_{z}(z)}[1 - \log D(G(z))]$.
The training process of the generator and the discriminator can be regraded as a two-player minimax game according to Eq.~\eqref{eq:L_GAN}.

\subsection{Semi-supervised Generative Adversarial Network}
\begin{figure}[htp]
\centering
\includegraphics[width=3.5in]{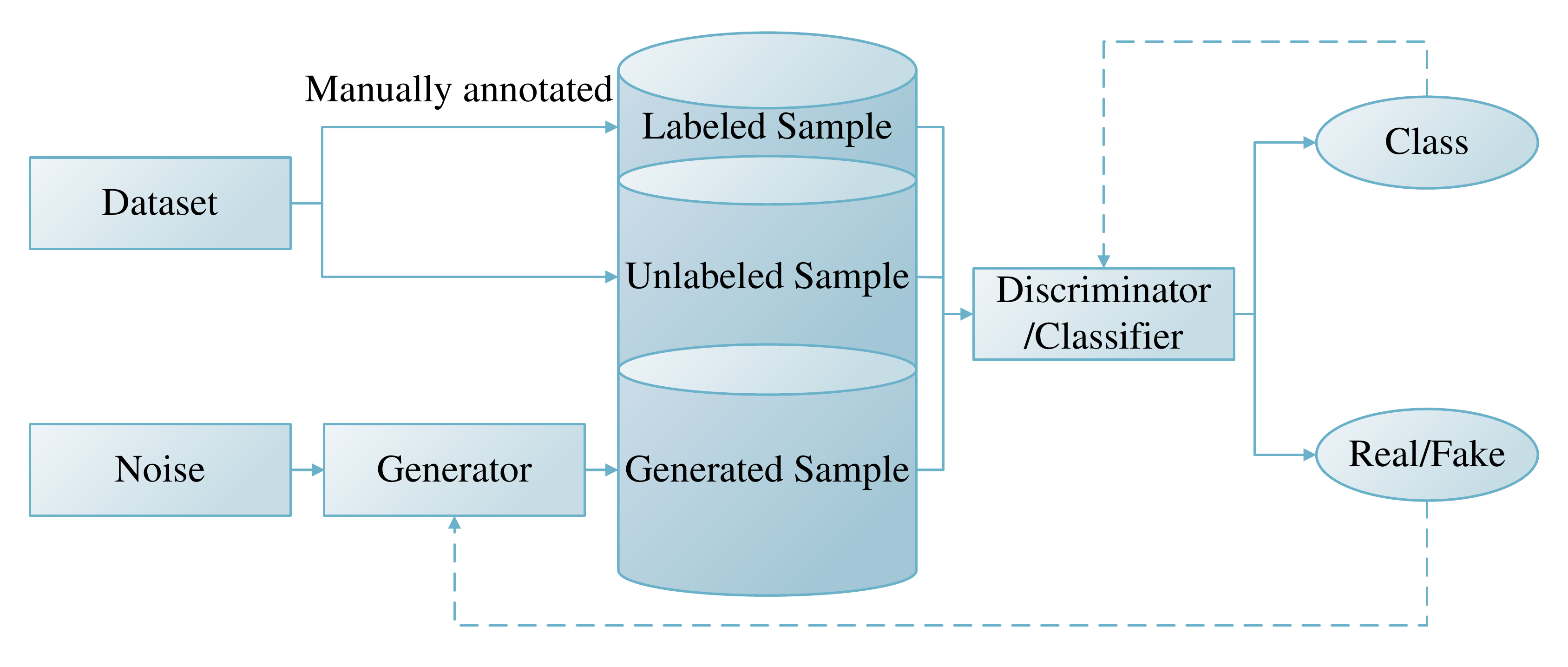}
\caption{The network structure of SSGAN.}
\label{fig:SSGAN}
\end{figure}

As an improved GAN, semi-supervised generative adversarial network~(SSGAN) has a network structure as shown in Fig.~\ref{fig:SSGAN}. 
Different from GAN, SSGAN can not only generate samples, but also can utilize label attribute to classify samples.

Specifically, SSGAN consists of two neural networks: a generator and a discriminator/classifier.
The training sample set $X$ of SSGAN consists of three parts: a small number of labeled samples $X_L$, a large number of unlabeled samples $X_U$ and generated samples $X_G$.
Let $\{1, 2, \dots, K\}$ represent the label space for classification and $p_D$ indicates the output of the discriminator.
Like~\cite{salimans2016improved}, the loss function of SSGAN consists of supervised loss and unsupervised loss:
\begin{equation}
\label{eq:SSGAN}
L_{\text{SSGAN}}=L_{\text{supervised}} + L_{\text{unsupervised}},
\end{equation}
where,
\begin{equation}
\label{eq:L_supervised}
L_{\text{supervised}} = -\mathbb{E}_{x,y\sim X_L}\log p_D(y|x,y\leq K),
\end{equation}
\begin{equation}
\begin{aligned}
\label{eq:L_unsupervised}
L_{\text{unsupervised}}=&-\mathbb{E}_{x\sim X_U}\log [1-p_D(y=K+1|x)]\\
& -\mathbb{E}_{x\sim X_G}\log p_D(y=K+1|x).
\end{aligned}
\end{equation}
It is seen from Eq.~\eqref{eq:L_supervised} that the supervised loss of SSGAN is consistent with the loss function of conventional fully supervised learning, of which purpose is to minimize the log conditional probability for labeled sample. In the unsupervised loss, according to Eq.~\eqref{eq:L_unsupervised}, the first term $-\mathbb{E}_{x\sim X_U}\log [1-p_D(y=K+1|x)]$ aims to minimize
the log conditional probability of
unlabeled sample being discriminated as
real, and the second term $-\mathbb{E}_{x\sim X_U}\log [1-p_D(y=K+1|x)]$ aims to minimize
the log probability of generated sample
being discriminated as fake.
Therefore, the unsupervised loss part of SSGAN is consistent with the discriminator loss part of the standard GAN.
The only difference is that the latter 
treats real samples as one class, while the former splits real samples into $K$ classes.

\subsection{Feature Matching Loss Function}
In GAN, standard adversarial loss has the following shortcomings.
(1) When the generator is updated with sigmoid cross-entropy loss function, it is easy to raise the problem of vanishing gradients for the samples that are on the correct side of the decision boundary, but are still far from the real data~\cite{mao2017least}.
(2) The generator may learn only part of modes of 
real-sample distribution, causing mode collapse.
(3) A perfect generator has no ability to improve the model’s generalization capacity.
Different from adversarial loss, feature matching loss aims to guide the generator to generate samples conducive to improving the generalization ability of SSGAN.
Feature matching loss function is defined as follows.
\begin{equation}
\label{eq:L_FM^l}
L_{\text{FM}}^{(l)}=\Vert \mathbb{E}_{x\sim X_G}\Phi^{(l)}(x)-\mathbb{E}_{x\sim X_U}\Phi^{(l)}(x) \Vert^{2},
\end{equation}
where $\Phi^{(l)}(x)$ represents the feature extracted from the  middle layer $l$ of the discriminator.
Obviously, samples that are far away
from the dataset manifold are meaningless for the classifier.
So the samples generated should be similar to the real data samples as a rule of thumb. 
Eq.~\eqref{eq:L_FM^l} represents that the L2 distance between generated samples and the real samples computed in the feature space of discriminator, hence guaranteeing the generated samples close to the real samples to some extent.
Feature matching loss is similar to perceptual loss computed in the feature space of a pre-trained VGG model~\cite{simonyan2014very,johnson2016perceptual}.

\section{Sea-land Clutter Semi-supervised Classification Based on WL-SSGAN}
\label{sec:Sea-land Clutter Semi-supervised Classification Based on WL-SSGAN}
In this section, we first describe the network structure and loss function of the proposed WL-SSGAN for sea-land clutter semi-supervised classification. Then, we introduce the training and algorithmic procedures of WL-SSGAN.

\subsection{Network Structure and Loss Function of WL-SSGAN}
\begin{figure}[htp]
\centering
\includegraphics[width=3.5in]{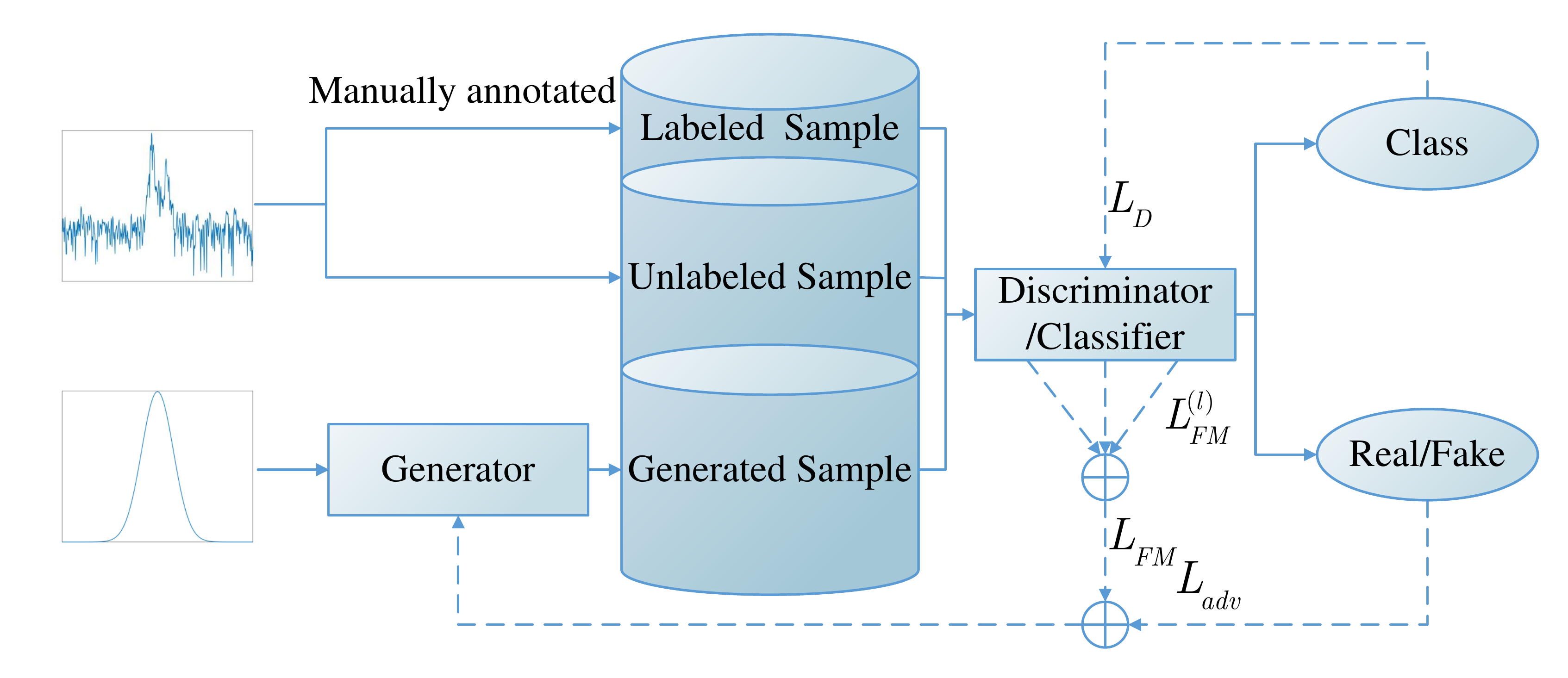}
\caption{The network structure of WL-SSGAN for sea-land clutter semi-supervised classification.}
\label{fig:WL-SSGAN}
\end{figure}

As the echo signals from OTHR, sea-land clutter samples have strong randomness in data distribution.
As a result, it is extremely difficult to train the generator of SSGAN with standard adversarial loss. On the contrary, the randomness enables feature matching loss to capture diverse sample features, which to some extent alleviates mode collapse for the generator and over-fitting for the discriminator.
To this end, we propose WL-SSGAN, which is a novel sea-land clutter semi-supervised classification framework. The network structure of WL-SSGAN is shown in Fig. \ref{fig:WL-SSGAN}, which is consistent with that of SSGAN. 
The discriminator loss $L_D$ of WL-SSGAN is consistent with that of SS-GAN:
\begin{equation}
\label{eq:L_D}
L_D=L_{\text{supervised}} + L_{\text{unsupervised}}.
\end{equation}

Different from SSGAN, WL-SSGAN improves the generator loss of SSGAN.
Specifically, by linearly weighting standard adversarial loss $L_{\text{adv}}$ and feature matching loss $L_{\text{FM}}^{(l)}$, a weighted loss $L_{\text{WL-SSGAN}}$ is proposed as follows, 
\begin{equation}
\begin{aligned}
\label{eq:L_WL-SSGAN}
L_{\text{WL-SSGAN}}&=\alpha L_{\text{adv}}+\beta L_{\text{FM}},
\end{aligned}
\end{equation}
where $\alpha+\beta=1$ and $\alpha, \beta \geq 0$. Additionally, in order to improve the generalization ability of WL-SSGAN, the signal features extracted from a single layer are not directly used for feature matching loss, but the multi-layer signal features are weighted to obtain joint feature matching loss.
Let $l_{\max}$ denote the total number of middle layers of the discriminator, $l_{\text{mul}}$ denote the set of selected feature layers, which has $2^{l_{\max}}-1$ combinations, and $Ch^{(l)}$ and $Le^{(l)}$ denote the channel number and signal length corresponding to the signal features of layer $l$, respectively.
The joint feature matching loss function is defined as follows,
\begin{equation}
\begin{aligned}
\label{eq:L_FM}
&L_{\text{FM}}=\sum\limits_{l\in l_{\text{mul}}}\frac{1}{2Ch^{(l)}Le^{(l)}}L_{\text{FM}}^{(l)}\\
&=\sum\limits_{l\in l_{\text{mul}}}\frac{1}{2Ch^{(l)}Le^{(l)}}\Vert \mathbb{E}_{x\sim X_G}\Phi^{(l)}(x)-\mathbb{E}_{x\sim X_U}\Phi^{(l)}(x) \Vert^{2}.
\end{aligned}
\end{equation}

For ease of understanding $L_{\text{WL-SSGAN}}$,
substitute $1-p_D(y=K+1\vert x)$ into  $L_{\text{unsupervised}}$ with $D(x)$. Then $L_{\text{unsupervised}}$ is converted to the standard GAN game-value:
\begin{equation}
\begin{aligned}
\label{eq:L_unsupervised2}
L_{\text{unsupervised}}=&-\mathbb{E}_{x\sim X_U}\log D(X)\\
&-\mathbb{E}_{x\sim X_G}\log [1-D(X)].
\end{aligned}
\end{equation}

Further, $L_{\text{WL-SSGAN}}$ can be rewritten as,
\begin{equation}
\begin{aligned}
\label{eq:L_WL-SSGAN2}
&L_{\text{WL-SSGAN}}\\
&=-\alpha(\mathbb{E}_{x\sim X_U}\log D(X)+\mathbb{E}_{x\sim X_G}\log [1-D(X)])\\
&+\beta \sum\limits_{l\in l_{\text{mul}}}\frac{1}{2Ch^{(l)}Le^{(l)}}\Vert \mathbb{E}_{x\sim X_G}\Phi^{(l)}(x)-\mathbb{E}_{x\sim X_U}\Phi^{(l)}(x) \Vert^{2}.
\end{aligned}
\end{equation}

It should be emphasized that the work in~\cite{springenberg2015unsupervised,odena2016semi,salimans2016improved} only took adversarial loss or feature matching loss as the generator loss of SSGAN, that is, they only emphasized the fidelity or diversity of the samples generated by the generator.
$L_{\text{WL-SSGAN}}$ combines the advantages of the above two loss functions. In order to obtain better semi-supervised classification performance, the proportion of weight factors $\alpha$ and $\beta$ can be controlled to balance the contribution of adversarial loss and joint feature matching loss to the training of WL-SSGAN.
In addition, $l_{\text{mul}}$ can also be selected to weight features of different layers to control the impact of different signal features on the classification performance of WL-SSGAN.

\subsection{Training Procedure of WL-SSGAN}
With the definitions of $L_D$ and $L_{\text{WL-SSGAN}}$, WL-SSGAN can be trained in a similar way to SSGAN. 
The detailed training procedure of WL-SSGAN is described in Algorithm \ref{alg:WL-SSGAN}.
If all the input signals have labels, only the parameters of classifier are updated for fully supervised learning according to the supervised loss part $L_{\text{supervised}}$ of $L_{D}$.
If the input contains both labeled and unlabeled signals, the parameters of the generator and the discriminator/classifier are alternately updated for semi-supervised learning according to the discriminator/classifier loss part $L_D$ and generator loss part $L_{\text{WL-SSGAN}}$.
\begin{algorithm}[!t]
\caption{Training of WL-SSGAN.}
\label{algorithm}
\begin{algorithmic}
\STATE \textbf{Initialize:} Generator, Discriminator/Classifier. Let $N$ be the total number of iterations. 
\FOR{$k = 1, \ldots, N$}
\STATE \textbf{Train Discriminator/Classifier}:
\STATE (1) Sample a mini-batch of a small number of labeled samples $X_L$;
\STATE (2) Sample a mini-batch of a prior noise distribution $p_z(z)$;
\STATE (3) Sample a mini-batch of a large number of unlabeled samples $X_U$;
\STATE (4) Update Discriminator/Classifier by $L_{D}$.
\STATE \textbf{Train Generator}:
\STATE (5) Repeat Steps (2)-(3);
\STATE (6) Update Generator by $L_{\text{WL-SSGAN}}$.
\ENDFOR
\end{algorithmic}
\label{alg:WL-SSGAN}
\end{algorithm}

\section{Experiment and Evaluation}
\label{sec:Experiment and Evaluation}
In this section, we first describe the sea-land clutter dataset, and then elaborate on the network design of WL-SSGAN.
Further, the network training details are introduced.
At last, we evaluate the proposed WL-SSGAN.

\subsection{Dataset}
\begin{figure}[!t]
\centering
\includegraphics[width=2.5in]{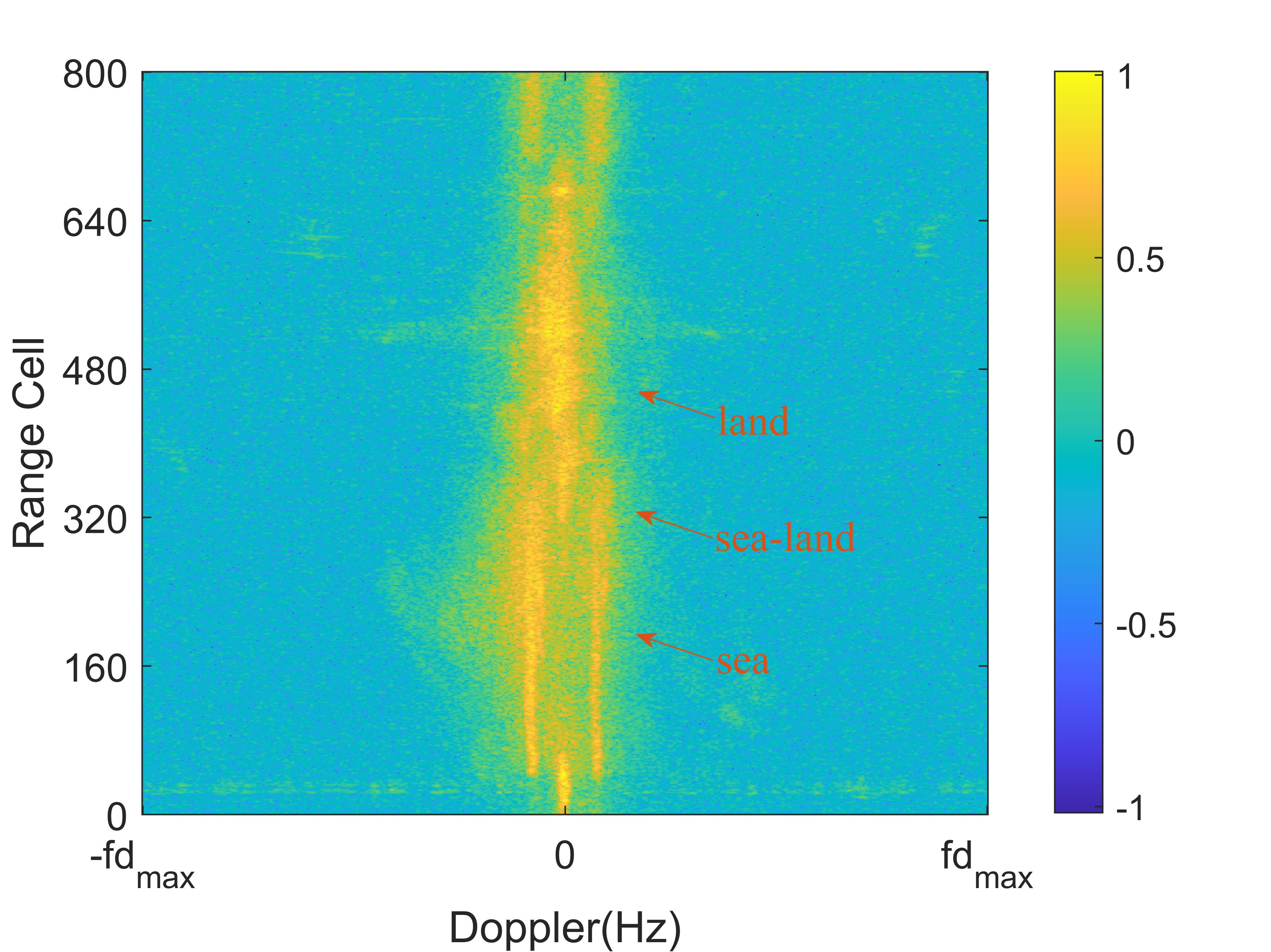}
\caption{The spectrum map of full range of a beam.}
\label{fig:spectrum}
\end{figure}

\begin{figure*}[!t]
\centering
\subfloat[]{\includegraphics[width=2in]{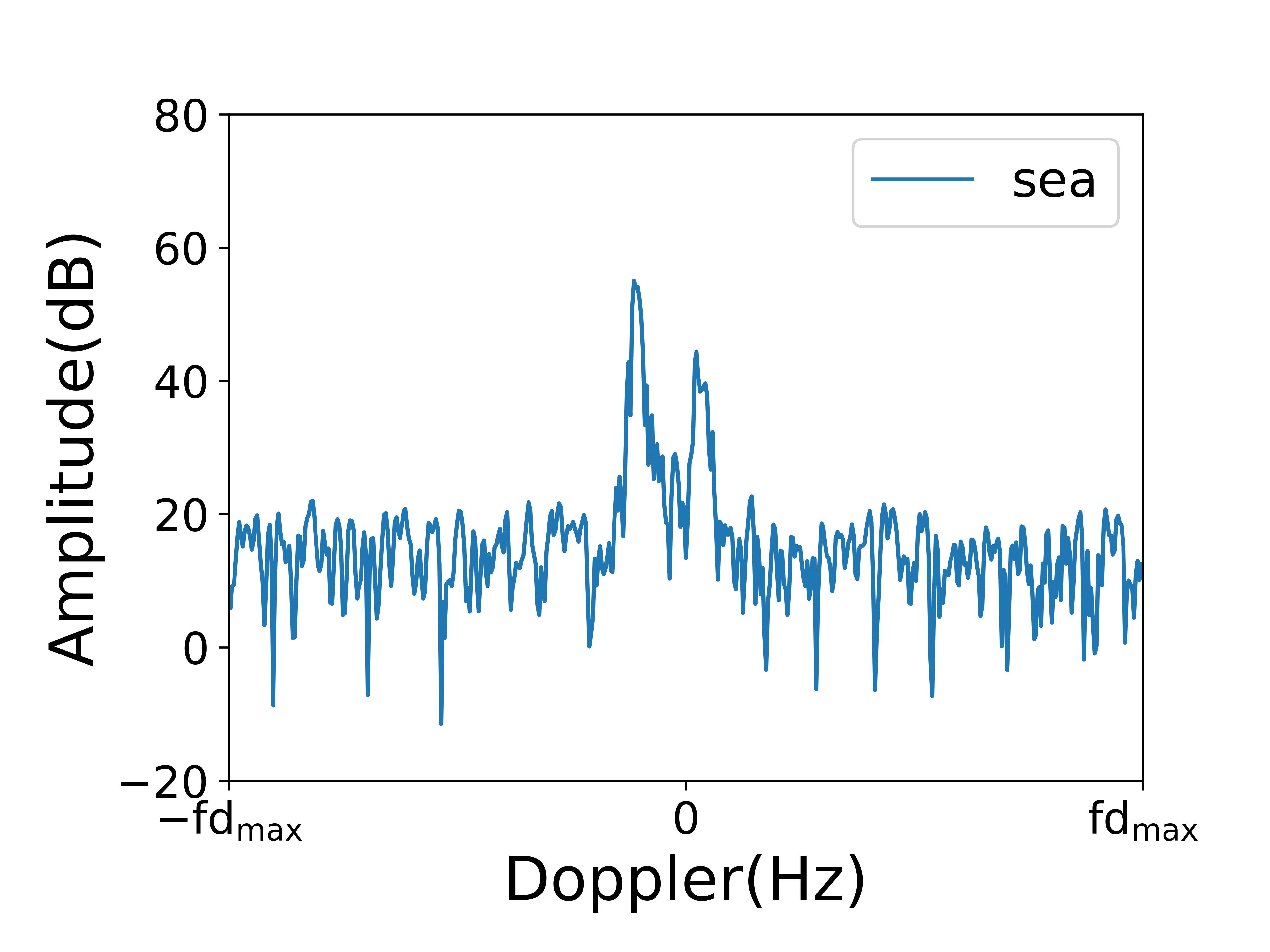}
\label{fig:sea_real}}
\hfil
\subfloat[]{\includegraphics[width=2in]{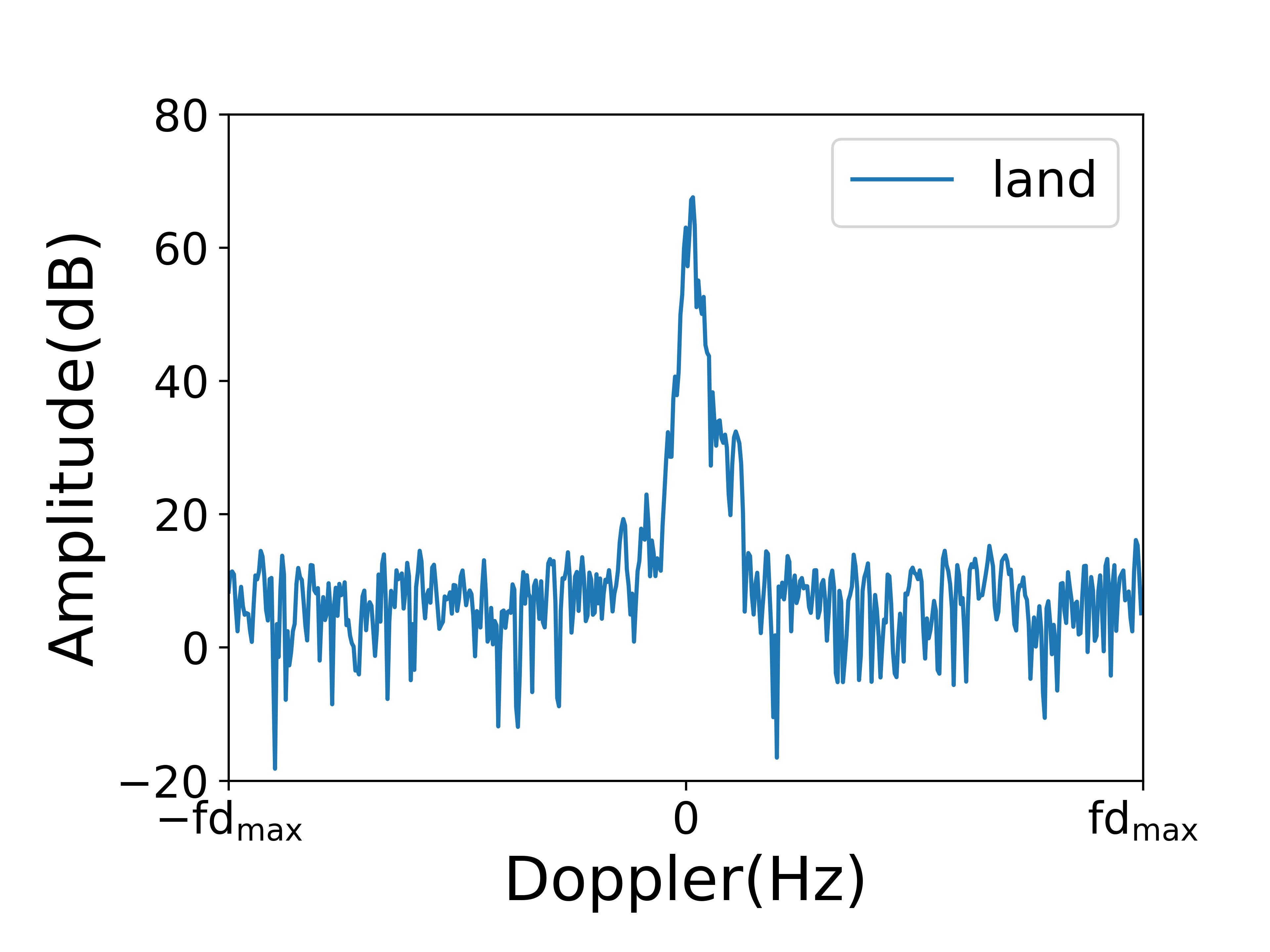}
\label{fig:land_real}}
\hfil
\subfloat[]{\includegraphics[width=2in]{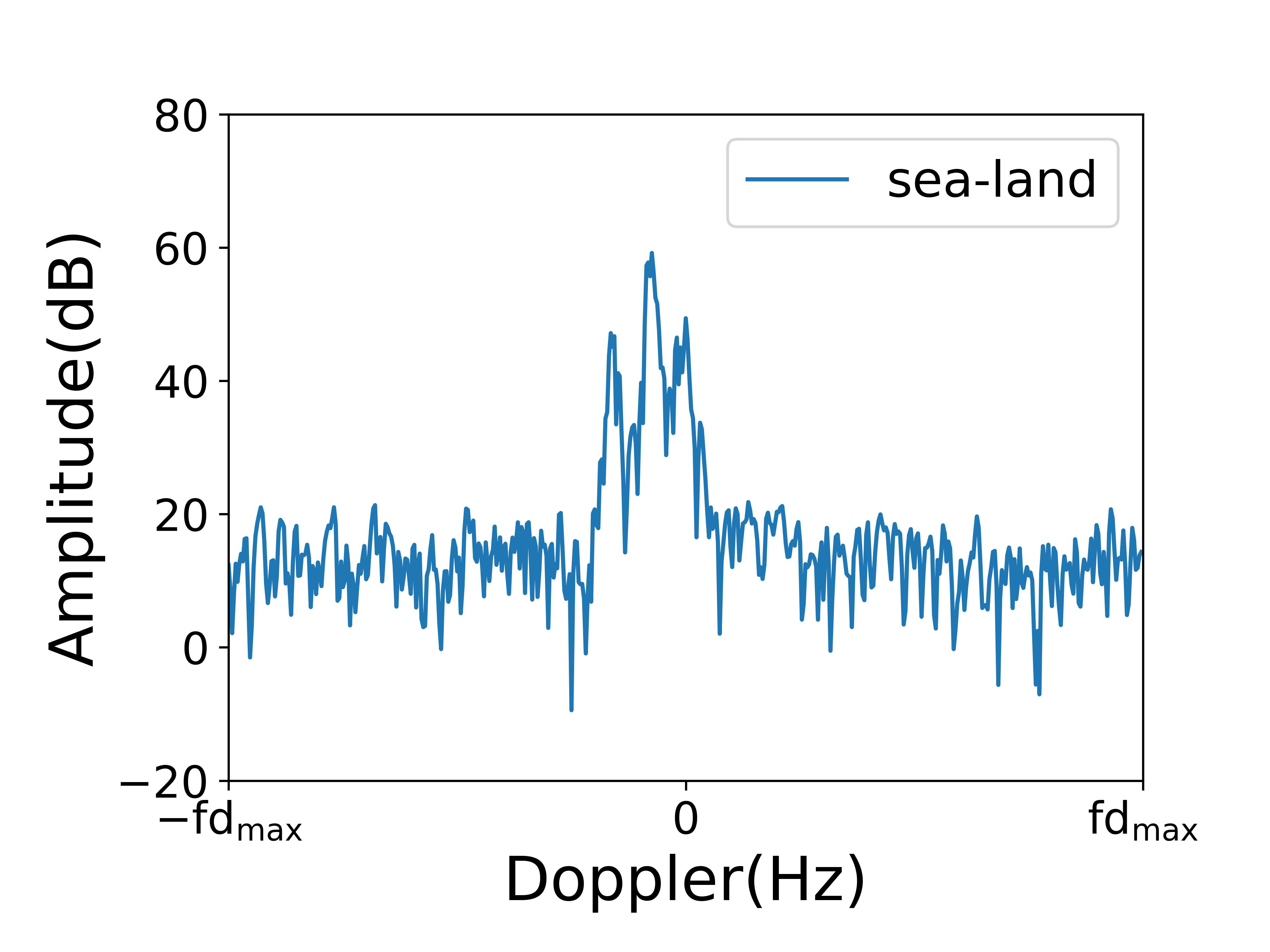}
\label{fig:sea-land_real}}
\caption{Sea-land clutter dataset. (a) An example of sea clutter. (b) An example of land clutter. (c) An example of sea-land boundary clutter.}
\label{fig:real sample}
\end{figure*}

To verify the semi-supervised classification performance of WL-SSGAN, we adopt sea-land clutter dataset as the benchmark dataset~\cite{zhang2022data}, which is the spectrum of clutter obtained by OTHR.
Fig.~\ref{fig:spectrum} shows the spectrum map of full range of a beam from OTHR. 
We take a collection of spectrum of a range-azimuth cell as sea-land clutter dataset, as shown in Fig.~\ref{fig:real sample}. 
Fig.~\ref{fig:real sample}\subref{fig:sea_real} is an example of sea clutter. 
The first-order Bragg peak of sea clutter is generated by the Bragg resonance scattering of high-frequency electromagnetic waves emitted by OTHR and ocean waves, showing double peaks symmetrical to zero frequency.
Fig.~\ref{fig:real sample}\subref{fig:land_real} is an example of land clutter. 
Since land has zero speed, it is observed that a single peak appears near zero frequency.
Fig.~\ref{fig:real sample}\subref{fig:sea-land_real} is an example of sea-land boundary clutter, which combines the features of sea clutter and land clutter, showing triple peaks. 
The detailed description of sea-land clutter dataset used to evaluate WL-SSGAN is shown in Table \ref{table:sea-land dataset}, in which training data is used for the training of WL-SSGAN and the evaluation of synthetic samples, and test data is only used for the evaluation of synthetic samples and do not participate in any training process.

\begin{table*}[!t]
\centering
\caption{Detailed Description of Sea-Land Clutter Dataset}
\label{table:sea-land dataset}
\begin{tabular}{cccccc}
\toprule
Class & Attribute & Label & Quantity & Training Quantity & Test Quantity\\
\hline
1 & sea & 0 & 1000& 700(70\%) & 300(30\%)\\
2 & land & 1 & 1000& 700(70\%) & 300(30\%)\\
3 & sea-land & 2 & 1000& 700(70\%) & 300(30\%)\\
\bottomrule
\end{tabular}
\end{table*}

Further, we randomly divide the original sea-land clutter dataset into labeled and unlabeled datasets, 
in which the test data is consistent with the original test data,
and only the label attribute of some samples from the original training data is removed. 
See Table~\ref{table:labeled datasets}.
For the convenience of description, we denote the total number of labeled samples as $n_{\text{lab}}$. 
The semi-supervised classification performance of WL-SSGAN is verified based on the above datasets.

\begin{table}[!t]
\centering
\caption{Sea-Land Clutter Training Data for Semi-supervised Classification}
\label{table:labeled datasets}
\setlength{\tabcolsep}{1mm}{
\begin{tabular}{ccccccccccc}
\toprule
 Attribute & \multicolumn{10}{c}{The number of labeled samples}\\
\hline
Sea & 10 & 20 & 30 & 40 & 50 & 100 & 200 & 300 & 400 & 500\\
Land & 10 & 20 & 30 & 40 & 50 & 100 & 200 & 300 & 400 & 500\\
Sea-Land & 10 & 20 & 30 & 40 & 50 & 100 & 200 & 300 & 400 & 500\\
Total & 30 & 60 & 90 & 120 & 150 & 300 & 600 & 900 & 1200 & 1500\\
\bottomrule
\end{tabular}
}
\end{table}

\subsection{Network Design}
\begin{figure}[!t]
\centering
\includegraphics[width=3.5in]{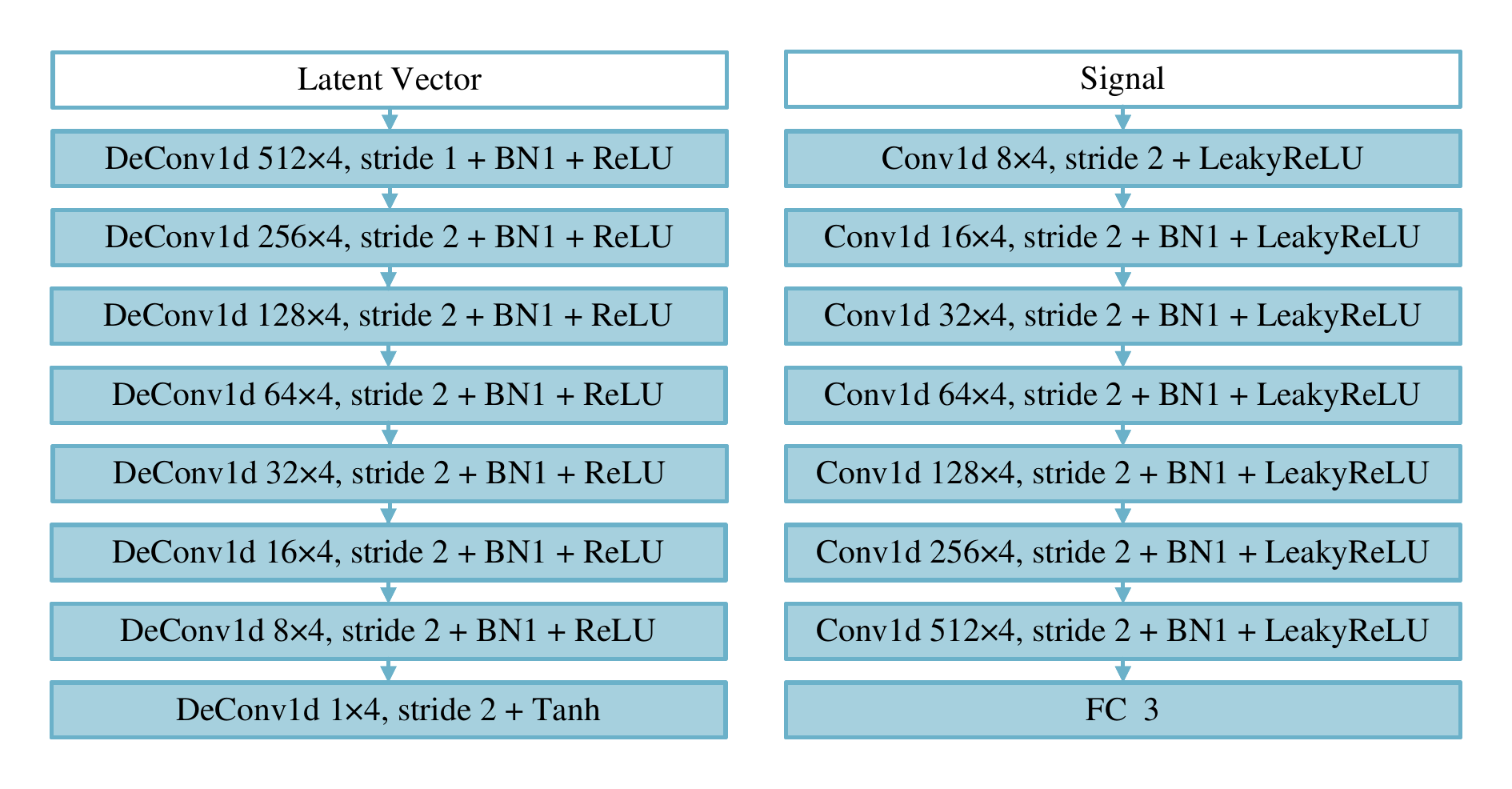}
\caption{The network structures of WL-SSGAN. Left:G and right:D.}
\label{fig:network}
\end{figure}

The network structure of WL-SSGAN for sea-land clutter semi-supervised classification is shown in Fig. \ref{fig:network}.

G consists of 8-layer neural network: eight 1-dimensional deconvolutional~(DeConv1d) layers followed by Batch Normalization~(BN1) and rectified linear unit~(ReLU)/hyperbolic tangent~(Tanh) activation function.
The input of G is a 100-dimensional latent vector~(Gaussian noise).
After layer 1, the number of channels becomes 512, and the dimension of the feature vector is four.
Then after layer 2-7, the number of channels is decreased by one-half, and the signal size is increased by two times per layer.
Finally, after layer 8, output a 512-dimensional synthetic sea-land clutter signal.

D consists of 8-layer neural network: seven 1-dimensional convolutional~(Conv1d) layers and one fully connected~(FC) layer.
All the Conv1d layers are followed by BN1 and LeakyReLU activation function with a slope of 0.2.
In addition, we add dropout with the probability of 0.5 after each Conv1d layer to prevent over-fitting.
The input of D is a $512$-dimensional sea-land clutter signal.
After layer 1, the number of channels is increased by eight times, and the signal size is decreased by one-half.
Then after layers 2-7, the number of channels is increased by two times per layer, and the signal size is decreased by one-half per layer.
Next, flatten the resulting feature vector.
Finally, output a 3-dimensional classification result.

\subsection{Network Training}
\label{sec:network training}
\begin{table}[!t]
\centering
\caption{Experiment Environment}
\label{table:enviroment}
\begin{tabular}{cc}
\toprule
Environment & Version\\
\hline
System & Windows 10 (64-bit)\\
GPU & NVIDIA GeForce RTX 3090\\
CUDA & 11.6\\
python & 3.9.0 (in Anaconda 4.11.0)\\
torch & 1.11.0\\
torchvision & 0.12.0\\
numpy & 1.22.3\\
matplotlib & 3.5.1\\
\bottomrule
\end{tabular}
\end{table}

\begin{table}[!t]
\centering
\caption{The Hyperparameter Configuration of WL-SSGAN for Sea-Land Clutter Semi-supervised Classification}
\label{table:paramater configuration}
\begin{tabular}{cc}
\toprule
Configuration & Default\\
\hline
Training Epoch & 1000\\
Batch Size & 64\\
Learning Rate & 0.0001\\
Cross Entropy Loss & -\\
Adma Optimizer & beta1:0.5, beta2:0.999\\
Data Normalization & [-1,1]\\
Weight Initialization & -\\
$\alpha, \beta$(weight factors) & $\alpha+\beta=1, \alpha,\beta\geq0$\\
$l_{\text{mul}}$(feature layers) & $[0, 2^{l_{\max}}-1]$\\
$n_{\text{lab}}$~(labeled samples) & -\\
\bottomrule
\end{tabular}
\end{table}

The experiment environment used in this work for sea-land clutter semi-supervised classification is shown in Table \ref{table:enviroment}.
See Table \ref{table:paramater configuration} for the training details and hyperparameter configuration of WL-SSGAN.

\begin{figure*}[!t]
\centering
\subfloat[]{\includegraphics[width=2.5in]{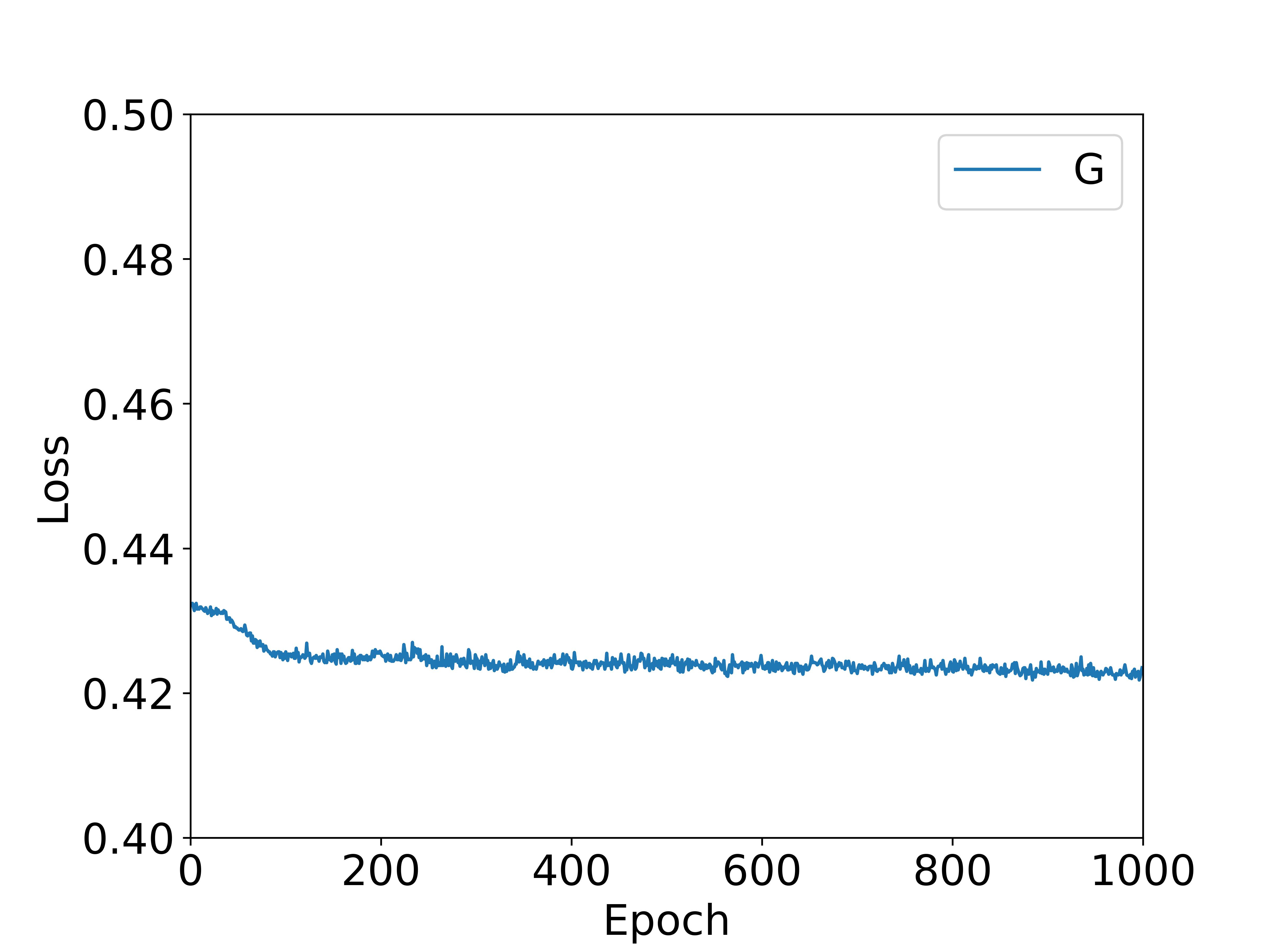}%
\label{fig:g_loss}}
\hfil
\subfloat[]{\includegraphics[width=2.5in]{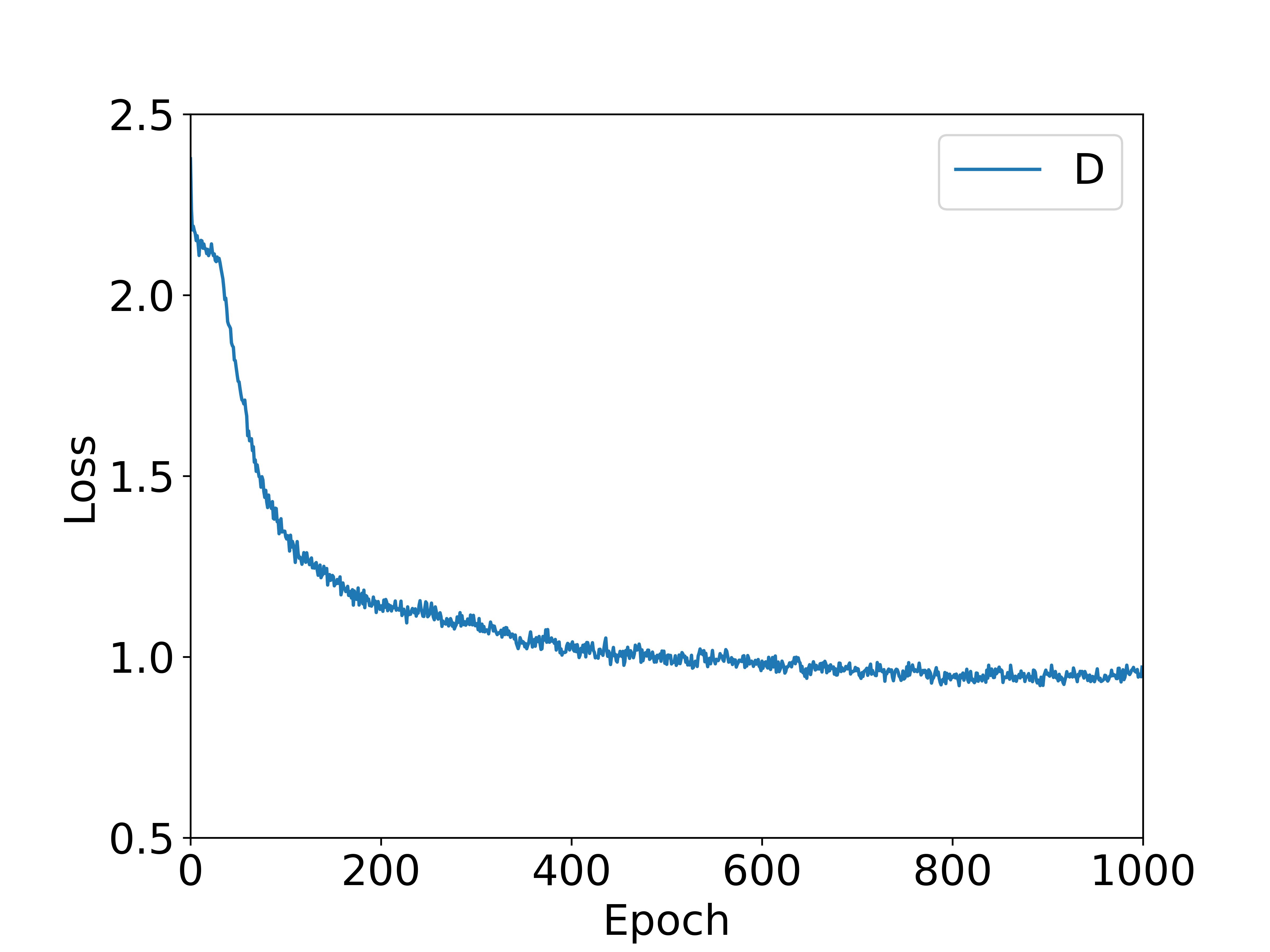}%
\label{fig:d_loss}}
\caption{Loss curves of WL-SSGAN for sea-land clutter synthesis~(with $\alpha=0.5$, $\beta=0.5$, $l_{\text{mul}}=[1,2,3,4,5,6,7]$ and $n_{\text{lab}}=2100$). (a) Loss curve for G. (b) Loss curve for D.}
\label{fig:gen_curve}
\end{figure*}

\begin{figure*}[!t]
\centering
\subfloat[]{\includegraphics[width=2.5in]{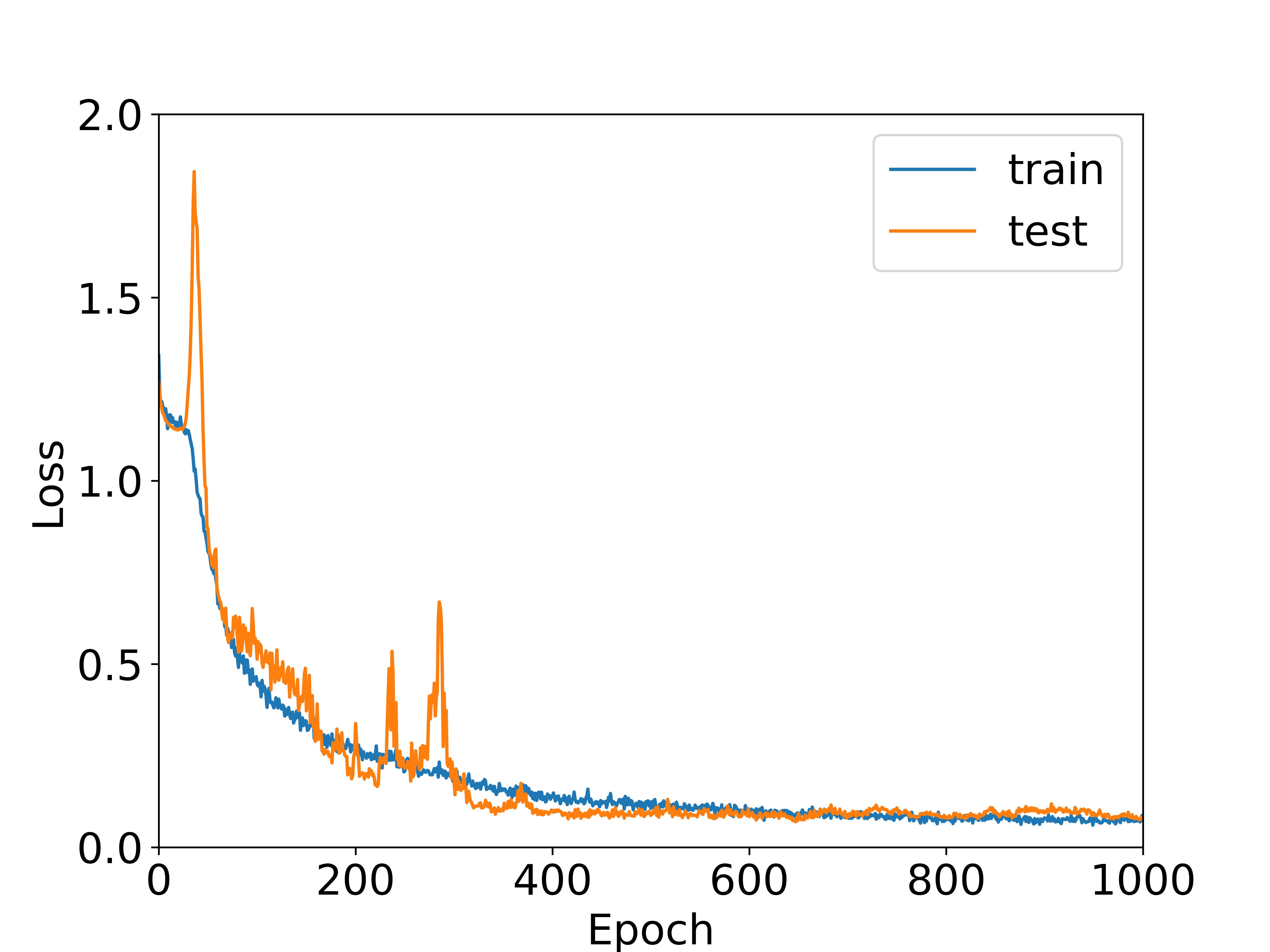}%
\label{fig:loss}}
\hfil
\subfloat[]{\includegraphics[width=2.5in]{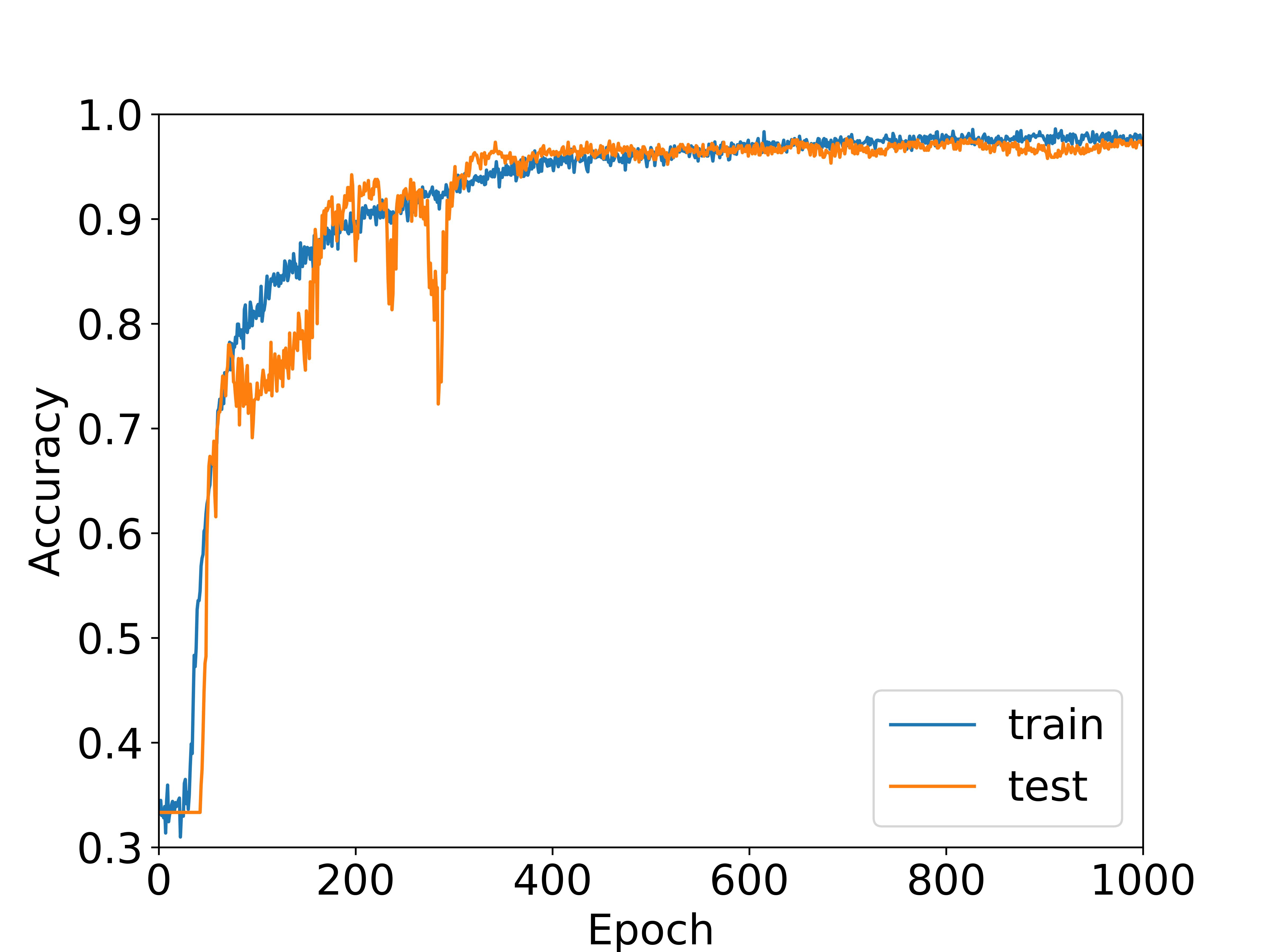}%
\label{fig:acc}}
\caption{Loss curve and accuracy curve of WL-SSGAN for sea-land clutter classification~(with $\alpha=0.5$, $\beta=0.5$, $l_{\text{mul}}=[1,2,3,4,5,6,7]$ and $n_{\text{lab}}=2100$). (a) Loss curve. (b) Accuracy curve.}
\label{fig:cla_curve}
\end{figure*}

Fig.~\ref{fig:gen_curve} and Fig.~\ref{fig:cla_curve} plot the loss curves and the classification accuracy of WL-SSGAN with $\alpha=0.5$, $\beta=0.5$, $l_{\text{mul}}=[1,2,3,4,5,6,7]$ and $n_{\text{lab}}=2100$, from which one can conclude that the training of WL-SSGAN is stable and converges rapidly.
From Fig.~\ref{fig:cla_curve}, one can observe that WL-SSGAN can achieve high-precision clutter classification.

\subsection{Network Evaluation}
Next, we consider the impact of the number of labeled samples $n_{\text{lab}}$, the proportion of $\alpha$ and $\beta$, and the selection of $l_{\text{mul}}$ on the classification performance of WL-SSGAN.
Additionally, the classification performance of WL-SSGAN is compared with that of a fully supervised classifier~(i.e. the discriminator of WL-SSGAN) trained with only a small number of labeled samples.

First, suppose that all the middle layer features of the discriminator contribute to generator loss, that is, $l_{\text{mul}}=\{1,2,3,4,5,6,7\}$. 
Thus, joint feature matching loss is fixed to evaluate the impact of the proportion of $\alpha$ and $\beta$ on the classification performance of WL-SSGAN.
Table \ref{table:alpha_beta} shows the average classification accuracy after WL-SSGAN reaches steady state.
\begin{table*}[!t]
\centering
\caption{The classification accuracy of WL-SSGAN under different values of $\alpha$ and $\beta$}
\label{table:alpha_beta}
\begin{tabular}{c|cccccccccc}
\hline
\hline
\diagbox{$(\alpha, \beta)$}{$\emph{n}_{\text{lab}}$} & 30 & 60 & 90 & 120 & 150 & 300 & 600 & 900 & 1200 & 1500\\
\hline
supervised & 0.7132 & 0.8046 & 0.8310 & 0.8521 & 0.8678 & 0.9039 & 0.9226 & 0.9372 & 0.9473 & 0.9627\\
\hline
$(0.0, 1.0)$ & 0.8663 & 0.8762 & 0.9517 & 0.9559 & 0.9618 & 0.9657 & 0.9678 & 0.9701 & 0.9790 & 0.9799\\
\hline
$(0.1, 0.9)$ & 0.8886 & 0.9129 & 0.9578 & 0.9645 & 0.9756 & 0.9764 & 0.9800 & 0.9841 & 0.9854 & 0.9901\\
\hline
$(0.2, 0.8)$ & 0.8679 & 0.9079 & 0.9618 & 0.9628 & 0.9643 & 0.9703 & 0.9715 & 0.9832 & 0.9836 & 0.9862\\
\hline
$(0.3, 0.7)$ & 0.8576 & 0.9070 & 0.9584 & 0.9677 & 0.9739 & 0.9768 & 0.9793 & 0.9827 & 0.9879 & 0.9888\\
\hline
$(0.4, 0.6)$ & 0.8883 & 0.9035 & 0.9537 & 0.9615 & 0.9719 & 0.9771 & 0.9782 & 0.9821 & 0.9835 & 0.9881\\
\hline
$(0.5, 0.5)$ & 0.8732 & 0.9023 & 0.9512 & 0.9561 & 0.9662 & 0.9720 & 0.9758 & 0.9812 & 0.9829 & 0.9857\\
\hline
$(0.6, 0.4)$ & 0.8636 & 0.9038 & 0.9514 & 0.9600 & 0.9626 & 0.9769 & 0.9810 & 0.9824 & 0.9836 & 0.9837\\
\hline
$(0.7, 0.3)$ & \textbf{0.8921} & \textbf{0.9280} & \textbf{0.9654} & \textbf{0.9700} & \textbf{0.9769} & \textbf{0.9801} & \textbf{0.9832} & \textbf{0.9868} & \textbf{0.9890} & \textbf{0.9908}\\
\hline
$(0.8, 0.2)$ & 0.8746 & 0.9012 & 0.9513 & 0.9634 & 0.9655 & 0.9723 & 0.9729 & 0.9804 & 0.9827 & 0.9879\\
\hline
$(0.9, 0.1)$ & 0.8601 & 0.9132 & 0.9487 & 0.9623 & 0.9713 & 0.9746 & 0.9769 & 0.9812 & 0.9833 & 0.9899\\
\hline
$(1.0, 0.0)$ & 0.8734 & 0.9179 & 0.9524 & 0.9637 & 0.9700 & 0.9734 & 0.9746 & 0.9832 & 0.9877 & 0.9882\\
\hline
\hline
\end{tabular}
\end{table*}

The following conclusions can be drawn from Table \ref{table:alpha_beta}.
(1) The classification performance of WL-SSGAN is better than that of fully supervised classifier, indicating that WL-SSGAN can extract potential features related to sea-land clutter classification from a large number of unlabeled samples.
Therefore, WL-SSGAN can improve the classification performance of fully supervised classifier trained with only a small number of labeled samples.
(2) With the decrease of $n_{\text{lab}}$, the  classification performance improvement achieved by WL-SSGAN is more and more obvious.
This is because when the number of training samples is small, fully supervised classifier is prone to fall into over-fitting.
(3) When the layers $l_{\text{mul}}$ of joint feature matching loss $L_{\text{FM}}$ is fixed, the proportion of $\alpha$ and $\beta$ obviously have an impact on classification accuracy.
The classification accuracy obtained only by using adversarial loss $L_{\text{adv}}$ or joint feature matching loss $L_{\text{FM}}$, that is, $(\alpha, \beta)=(1.0, 0.0)$ or $(\alpha, \beta)=(0.0, 1.0)$, is not the best.
The best classification accuracy is obtained when $(\alpha, \beta)=(0.7, 0.3)$.
It is concluded that the proposed weighted loss $L_{\text{WL-SSGAN}}$ is superior to both $L_{\text{adv}}$ and $L_{\text{FM}}$.

Second, suppose that adversarial loss and joint feature matching loss have the same contribution to generator loss, that is, $(\alpha, \beta)=(0.5, 0.5)$.
Thus, the contribution of adversarial loss and joint feature matching loss to WL-SSGAN is fixed to evaluate the selection of $l_{\text{mul}}$ on the classification performance of WL-SSGAN.
Table \ref{table:l_mul} shows the average classification accuracy after WL-SSGAN reaches steady state.
\begin{table*}[!t]
\centering
\caption{The classification accuracy of WL-SSGAN under different value of $l_{\rm{mul}}$}
\label{table:l_mul}
\begin{tabular}{c|cccccccccc}
\hline
\hline
\diagbox{$l_{\text{mul}}$}{$\emph{n}_{\text{lab}}$} & 30 & 60 & 90 & 120 & 150 & 300 & 600 & 900 & 1200 & 1500\\
\hline
supervised & 0.7132 & 0.8046 & 0.8310 & 0.8521 & 0.8678 & 0.9039 & 0.9226 & 0.9372 & 0.9473 & 0.9627\\
\hline
\makecell[l]{$\{1\}$} & 0.8621 & 0.9030 & 0.9500 & 0.9535 & 0.9626 & 0.9702 & 0.9714 & 0.9787 & 0.9813 & 0.9844\\
\hline
\makecell[l]{$\{1, 2\}$} & 0.8567 & 0.8985 & 0.9521 & 0.9543 & 0.9676 & 0.9687 & 0.9721 & 0.9830 & 0.9856 & 0.9875\\
\hline
\makecell[l]{$\{1, 2, 3\}$} & 0.8765 & 0.9063 & 0.9573 & 0.9585 & 0.9662 & 0.9670 & 0.9743 & 0.9831 & 0.9854 & 0.9867\\
\hline
\makecell[l]{$\{1, 2, 3, 4\}$} & 0.8611 & 0.9163 & 0.9551 & 0.9610 & 0.9623 & 0.9731 & 0.9745 & 0.9833 & 0.9841 & 0.9852\\
\hline
\makecell[l]{$\{1, 2, 3, 4, 5\}$} & 0.8673 & 0.9057 & 0.9568 & 0.9573 & 0.9651 & 0.9710 & 0.9764 & 0.9832 & 0.9841 & 0.9855\\
\hline
\makecell[l]{$\{1, 2, 3, 4, 5, 6\}$} & \textbf{0.8960} & \textbf{0.9209} & \textbf{0.9603} & \textbf{0.9643} & \textbf{0.9701} & \textbf{0.9746} & \textbf{0.9782} & \textbf{0.9847} & \textbf{0.9868} & \textbf{0.9903}\\
\hline
\makecell[l]{$\{1, 2, 3, 4, 5, 6, 7\}$} & 0.8921 & 0.9020 & 0.9553 & 0.9565 & 0.9666 & 0.9721 & 0.9732 & 0.9821 & 0.9833 & 0.9864\\
\hline
\hline
\end{tabular}
\end{table*}

The following conclusions can be drawn from Table \ref{table:l_mul}.
(1) When the proportion of weight factors $\alpha$ and $\beta$ of adversarial loss $L_{\text{adv}}$ and joint feature matching loss $L_{\text{FM}}$ are fixed, the selection of $l_{\text{mul}}$ obviously has an impact on classification accuracy.
The classification accuracy obtained by using the feature matching loss $L_{\text{FM}}^{(1)}$ with single layer or the joint feature matching loss with all layers $L_{\text{FM}}^{(1)-(7)}$, that is, $l_{\text{mul}}=\{1\}$ or $l_{\text{mul}}=\{1, 2, 3, 4, 5, 6, 7\}$, is not the best.
The best classification accuracy is obtained when $l_{\text{mul}}=\{1, 2, 3, 4, 5, 6\}$.
This indicates that:
(1) The proposed joint feature matching loss $L_{\text{FM}}$ is superior to conventional feature matching loss $L_{\text{FM}}^{(1)}$.  
(2) Not all the middle layer features of discriminator are weighted to obtain the best semi-supervised classification performance.
On the contrary, the features extracted from some layers may inhibit classification performance.
It is concluded that the best classification performance depends on the appropriate selection of feature layers.

Further, we compare WL-SSGAN with traditional full supervised classification methods~\cite{zhang2022data}. See Table~\ref{table:classification model}.
It is seen that the classification accuracy of WL-SSGAN using only 90 labeled samples is greater than those of Random Forest, KNN, Logistic Regression and SVM using all labeled samples.
The WL-SSGAN using only 1200 labeled samples has the best classification accuracy.

\begin{table}[!t]
\centering
\caption{Classification Accuracy of Different Classifiers on Sea-Land Clutter Dataset}
\label{table:classification model}
\begin{tabular}{cc}
\toprule
Classifiers & Accuracy\\
\hline
Random Forest & 94.18\%~(all labeled) \\
KNN & 94.76\%~(all labeled) \\
Logistic Regression & 95.36\%~(all labeled) \\
SVM & 95.82\%~(all labeled) \\
\textbf{WL-SSGAN~(Ours)} & \textbf{96.03\%~(90 labeled)}\\
FCN & 97.93\%~(all labeled) \\
\textbf{WL-SSGAN~(Ours)} & \textbf{98.01\%~(300 labeled)}\\
ResNet18 & 98.87\%~(all labeled) \\
\textbf{WL-SSGAN~(Ours)} & \textbf{98.90\%~(1200 labeled)}\\
\bottomrule
\end{tabular}
\end{table}

\begin{figure}[!t]
\centering
\includegraphics[width=3in]{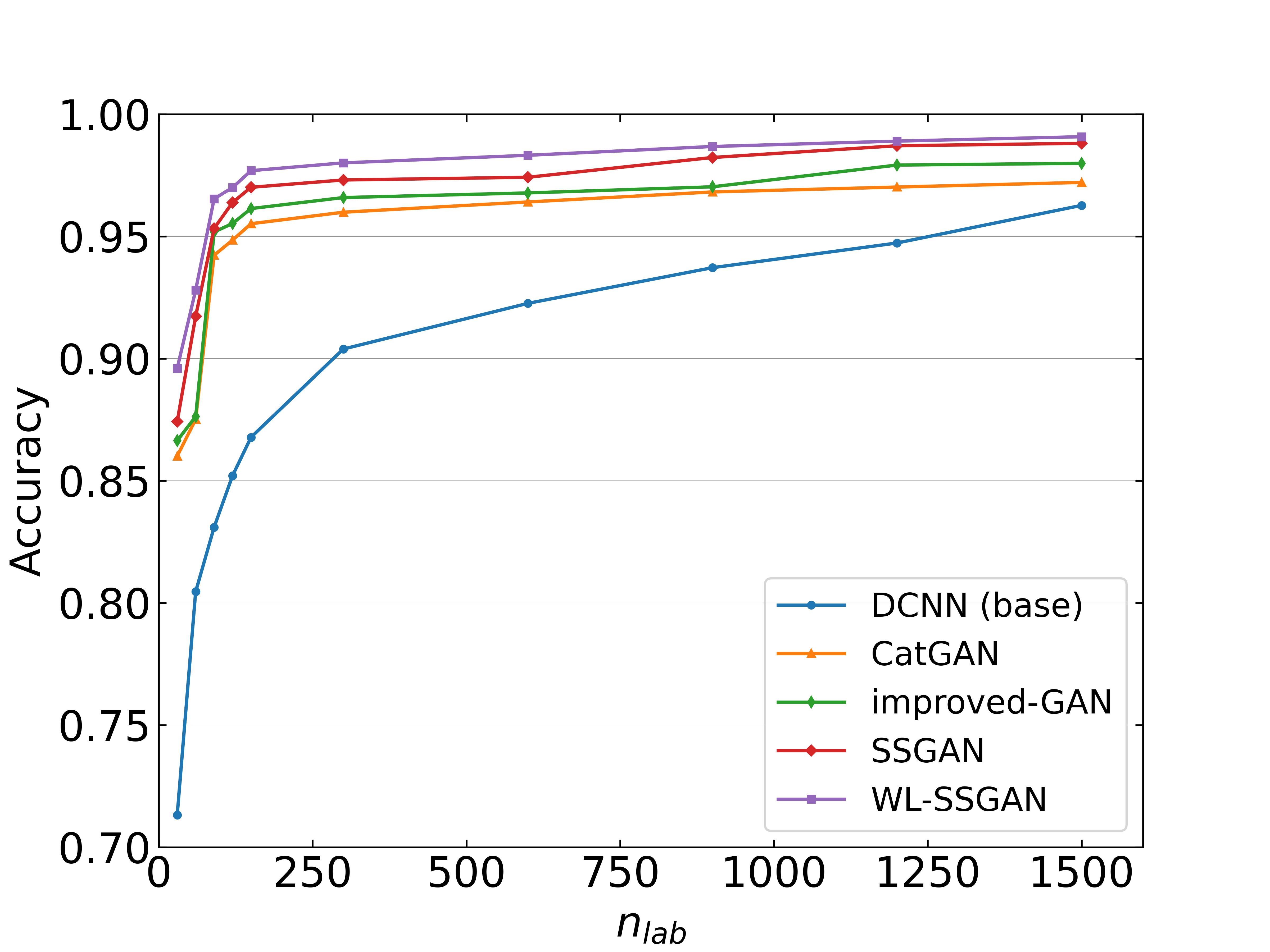}
\caption{Classification accuracy of different semi-supervised classifiers on different numbers of labeled sea-land clutter samples.}
\label{fig:semi_acc}
\end{figure}

Besides, we compare the classification accuracy of the different semi-supervised classifiers on different numbers of labeled sea-land clutter samples, in which the network structure and hyperparameter configuration of all methods are consistent.
See Fig.~\ref{fig:semi_acc}.
It is seen that WL-SSGAN has advantages over traditional methods in semi-supervised classification. 

\begin{figure*}[!t]
\centering
\subfloat[]{\includegraphics[width=2in]{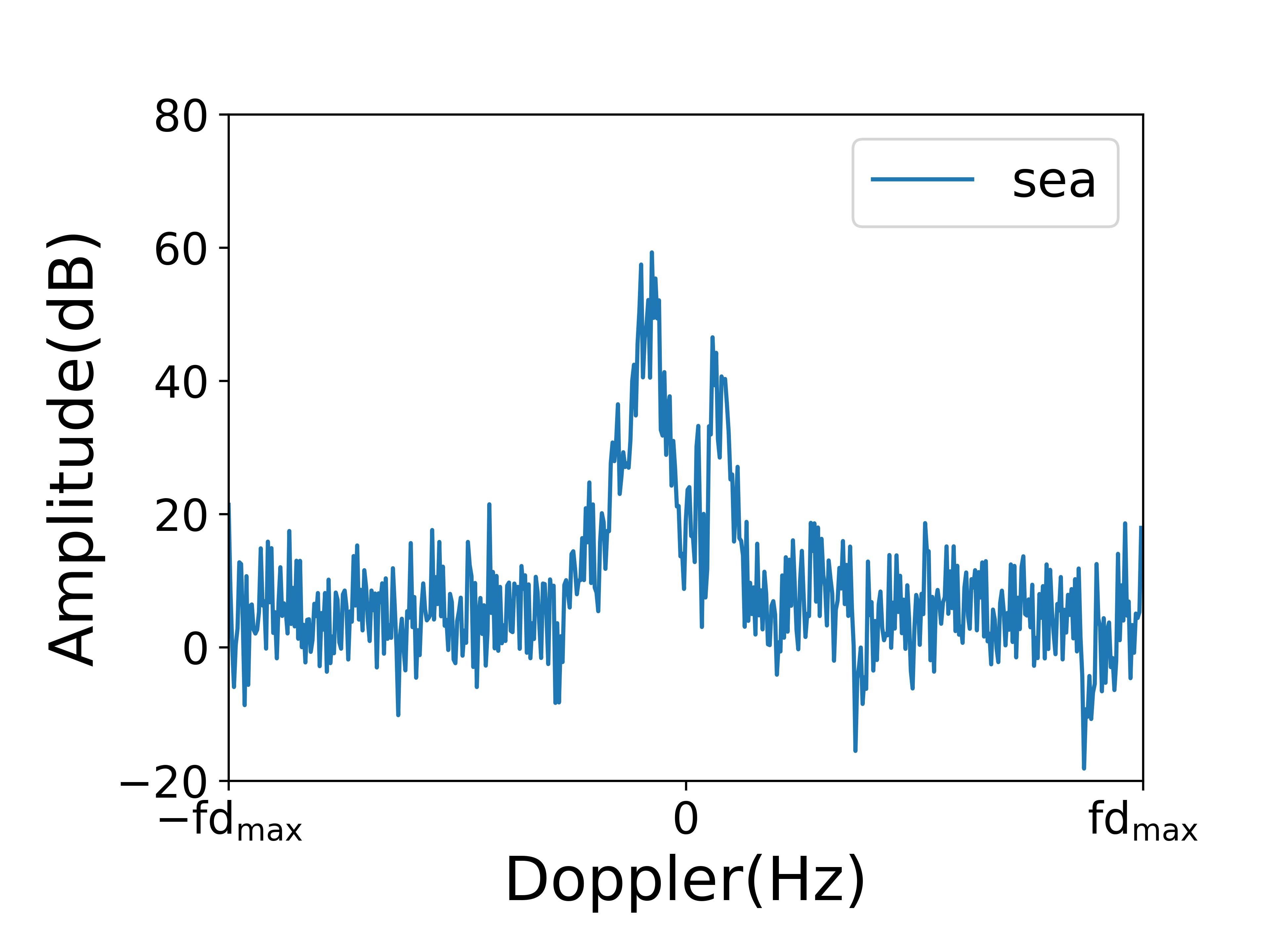}
\label{fig:sea_gen}}
\hfil
\subfloat[]{\includegraphics[width=2in]{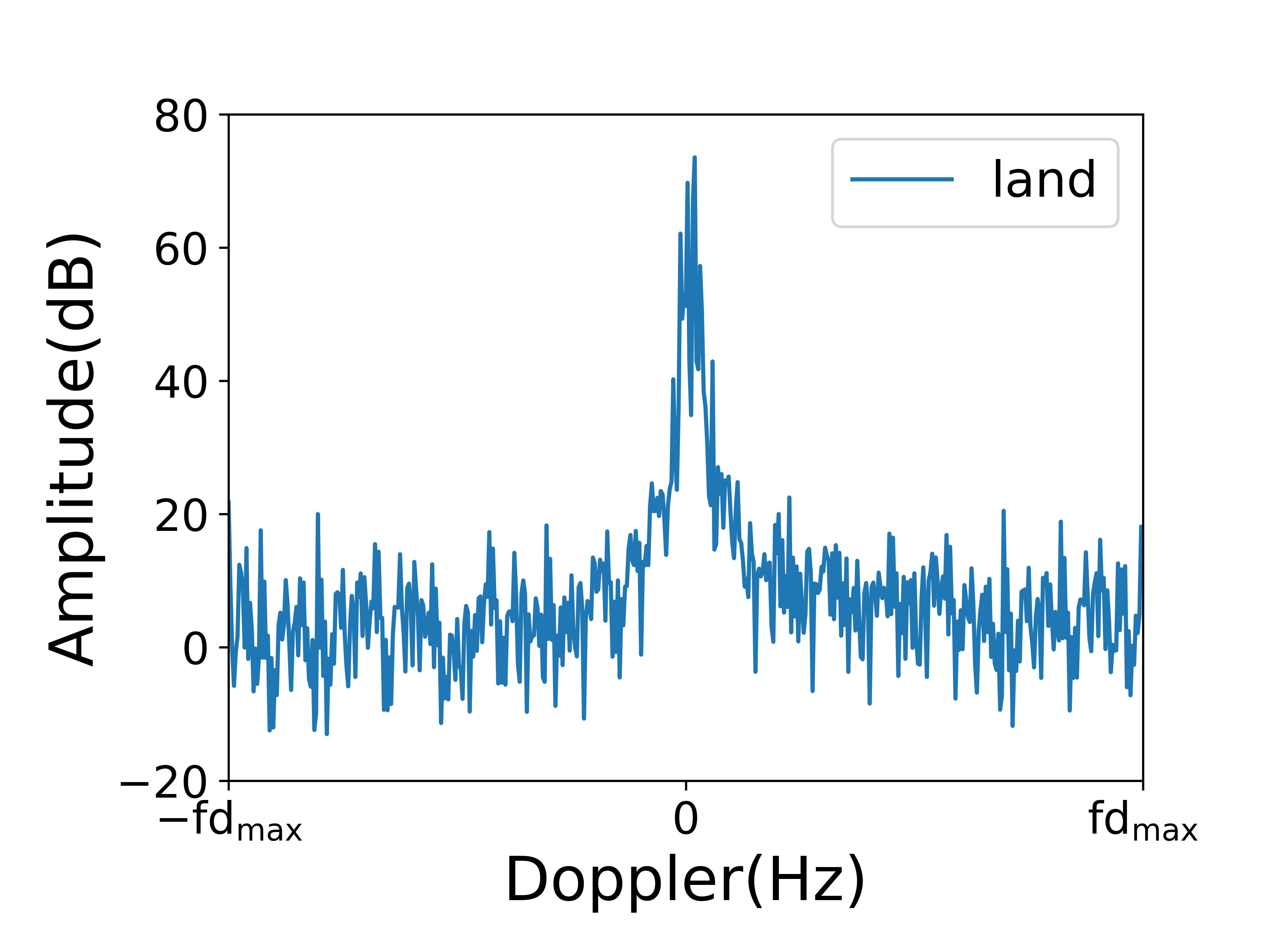}
\label{fig:land_gen}}
\hfil
\subfloat[]{\includegraphics[width=2in]{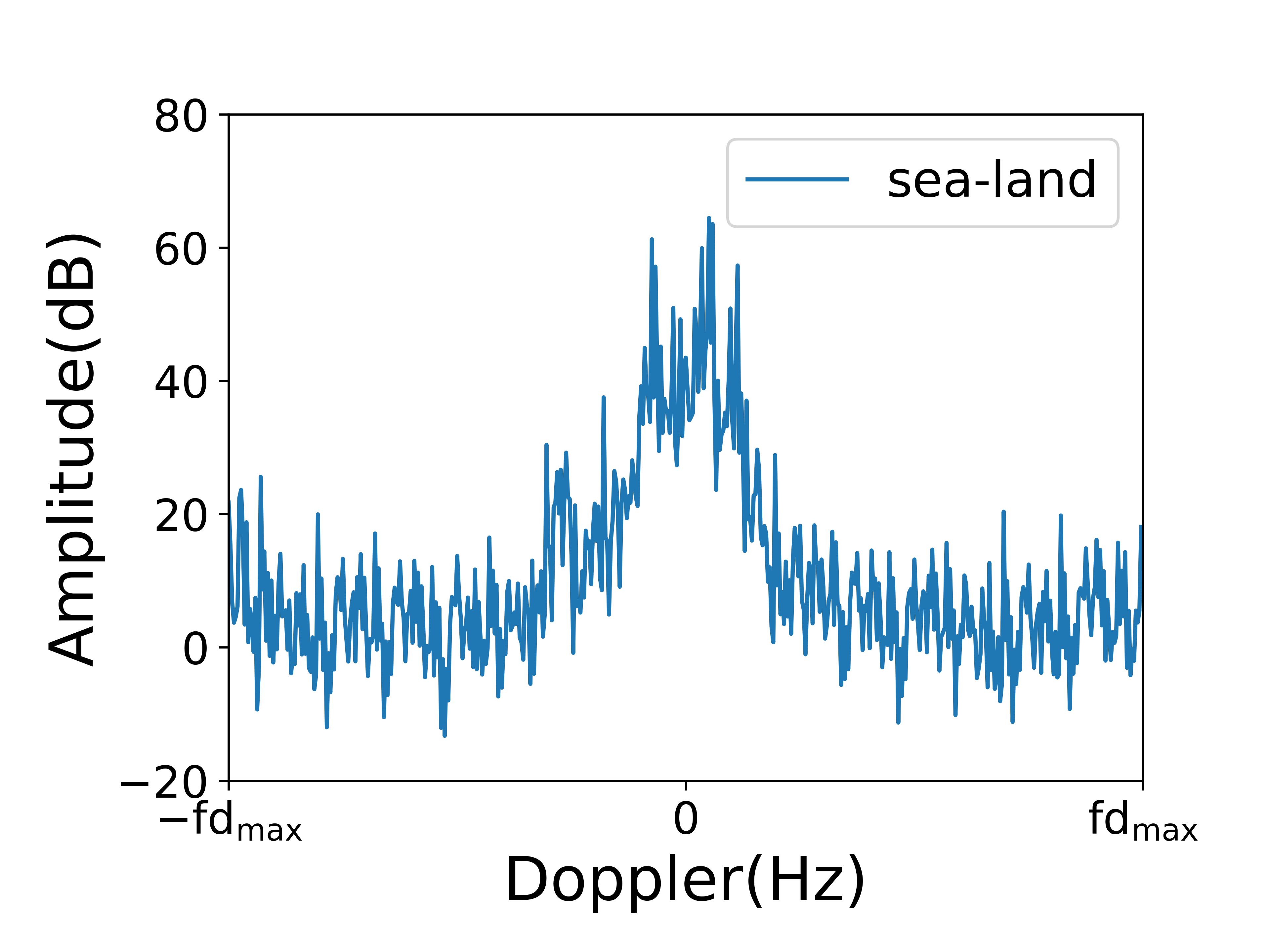}
\label{fig:sea-land_gen}}
\caption{The sea-land clutter samples synthesized by WL-SSGAN~(with $\alpha=0.5$, $\beta=0.5$, $l_{\text{mul}}=[1,2,3,4,5,6,7]$ and $n_{\text{lab}}=2100$). (a) sea clutter. (b) land clutter. (c) sea-land boundary clutter.}
\label{fig:gen sample}
\end{figure*}

Finally, we evaluate the sample synthesis performance of WL-SSGAN. We use trained WL-SSGAN in Section \ref{sec:network training} to synthesize 2100 sea-land clutter samples. See Fig.~\ref{fig:gen sample} for some examples of synthetic samples.
Visually, most of synthetic samples are consistent with the real samples.
Note that WL-SSGAN cannot synthesize samples with specific categories, while we observe an abundance of sea clutter samples, land clutter samples and sea-land boundary clutter samples from the synthetic samples.
This illustrates that WL-SSGAN also has good sample synthesis ability.
Further, we evaluate the quality of the synthesized sea-land clutter in terms of the statistical evaluation metrics including absolute distance~(AD), cosine similarity~(CS) and Pearson correlation coefficient~(PCC)~\cite{zhang2022data}. 
See Table~\ref{table:synthesis evaluation} for the evaluation results.
It is seen that the sample synthesis performance of WL-SSGAN outperforms those of DCGAN, WGAN-GP, LS-GAN, VAE-GAN and CGAN.
WL-SSGAN obtains competitive results compared to that of AC-GAN.
The sample synthesis ability of WL-SSGAN is slightly lower than that of AC-VAEGAN.
Note that WL-SSGAN is more focused on semi-supervised classification.

\begin{table}[!t]
\centering
\caption{The Evaluation Results of Synthetic Sea-Land Clutter Samples}
\label{table:synthesis evaluation}
\centering
\setlength{\tabcolsep}{1mm}{
\begin{tabular}{cccc}
\toprule
GAN Methods & \multicolumn{3}{c}{Statistical Evaluation} \\
   \cmidrule(r){2-4} 
    & AD & CS& PCC \\
\hline
DCGAN & 0.0313 & 0.7548 & 0.7534 \\
WGAN-GP & 0.0283 & 0.7671 & 0.7483 \\
LSGAN & 0.0379 & 0.7106 & 0.7521 \\
VAEGAN & 0.0244 & 0.7337 & 0.7518 \\
CGAN & 0.0221 & 0.7882 & 0.7458 \\
AC-GAN & 0.0161 & 0.8036 & 0.7649 \\
\textbf{AC-VAEGAN} & \textbf{0.0160} & \textbf{0.8988} & \textbf{0.7897} \\
WL-SSGAN(Ours) & 0.0173 & 0.8672 & 0.7558 \\
\bottomrule
\end{tabular}
}
\end{table}

\section{Conclusions and Future Work}
\label{sec:Conclusions and Future Work}
A novel sea-land clutter semi-supervised classification framework for OTHR was proposed, namely WL-SSGAN.
Considering the randomness of sea-land clutter samples, joint feature matching loss was proposed by weighting multi-layer feature matching loss. Further, adversarial loss and joint feature matching loss were weighted to alleviate the effect of sea-land clutter samples with obvious noise on semi-supervised classification performance.
The experimental results showed that WL-SSGAN can achieve semi-supervised classification of sea-land clutter, and the proposed weighted loss is superior to both the adversarial loss and the feature matching loss.

In future work, there are two issues that need to be addressed:
(1) While WL-SSGAN improves classification performance by utilizing a large number of unlabeled sea-land clutter samples, the computational cost increases.
Therefore, it is desirable to improve the semi-supervised classification framework of WL-SSGAN to take into account both the classification performance and the computational cost.
(2) The proposed weighted loss was determined by the selection of $l_{\text{mul}}$ and the proportion of $\alpha$ and $\beta$.
The values of $l_{\text{mul}}$, $\alpha$ and $\beta$ are considered as hyperparameters in the experiment.
We intend to design the optimal selection of $l_{\text{mul}}$ and the optimal proportion of $\alpha$ and $\beta$.
Furthermore, we expect to find a parameter adaptive optimization scheme to automatically seek the values of $l_{\text{mul}}$, $\alpha$ and $\beta$ with the best semi-supervised classification performance.

\bibliographystyle{IEEEtran}
\bibliography{references}

\end{document}